\pdfoutput=1
\documentclass[11pt]{article}

\usepackage{EMNLP2023}

\usepackage{times}
\usepackage{latexsym}
\usepackage[T1]{fontenc}
\usepackage[utf8]{inputenc}
\usepackage{mathtools}
\usepackage{microtype}

\usepackage{inconsolata}
\usepackage{graphicx}
\usepackage[inkscapeformat=png]{svg}
\usepackage{svg}
\usepackage{graphicx}
\usepackage{multicol}
\usepackage{graphicx} % Required for \resizebox
\usepackage{tabularx} % Required for the tabularx environment
\usepackage{caption} % Required for \caption command

\usepackage{inconsolata}
\usepackage[inkscapeformat=png]{svg}
\usepackage{svg}
\usepackage{natbib}
\usepackage{amsmath}
\usepackage[utf8]{inputenc}
\usepackage{mathtools}
\usepackage{colortbl}
\usepackage{microtype}
\usepackage{tikz}

\usepackage{amsfonts}
\usepackage{tabularx} 
\usepackage{booktabs} 
\usepackage{cite} 

\usepackage{inconsolata}
\usepackage{graphicx}

\usepackage{hyperref}

\usepackage{xparse}
\newcounter{chatlinenum}

\makeatletter
\newif\ifacl@finalcopy
\acl@finalcopytrue
\makeatother

\tikzset{
  chatstyle/.style={
    text width=0.85\textwidth,
    rounded corners=5pt, % Slightly rounder corners
    align=left,
    font=\footnotesize, % Smaller text size
    inner sep=1pt, % Padding around the text    
  }
}

\definecolor{mygreen}{HTML}{a8caea}

\NewDocumentCommand{\chatline}{m}{%
    \stepcounter{chatlinenum}%
    \par\noindent% Ensure there is no indentation and start a new paragraph
    \ifodd\thechatlinenum
        \tikz\node[fill=lightgray,chatstyle]{
            \begin{minipage}[t]{.05\textwidth}
                \includegraphics[width=\linewidth]{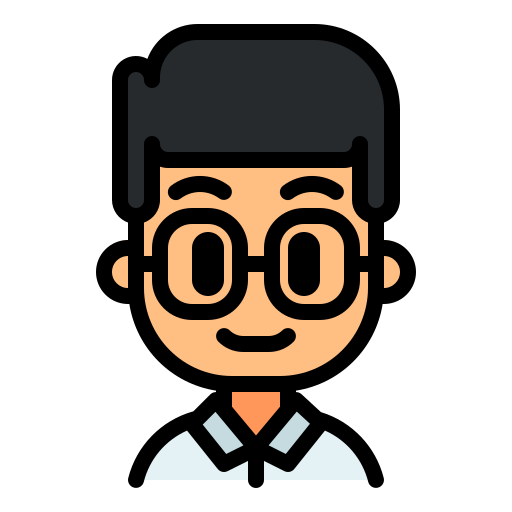}
            \end{minipage}%
            \begin{minipage}[t]{.88\textwidth}
                #1
            \end{minipage}
        };
    \else
        \hfill\tikz\node[fill=mygreen,chatstyle,align=right]{
            \begin{minipage}[t]{.88\textwidth}
                #1
            \end{minipage}%
            \begin{minipage}[t]{.05\textwidth}
                \includegraphics[width=\linewidth]{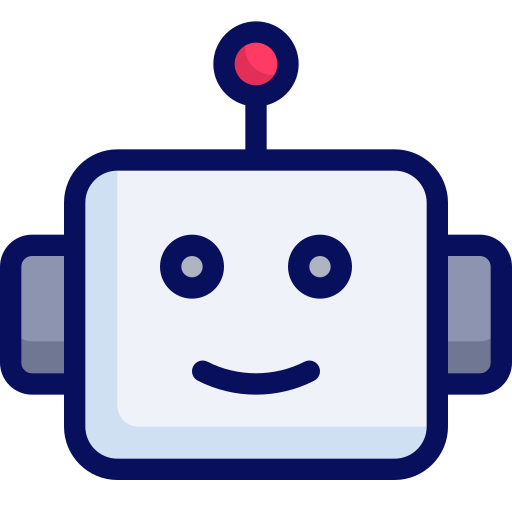}
            \end{minipage}
        };
    \fi
    \par% End the paragraph to allow the next content to be placed correctly
    \smallskip% Add a small vertical space after the chat bubble
}

\NewDocumentEnvironment{newchat}{}{%
    \setcounter{chatlinenum}{0}
}

\title{\includegraphics[height=0.8cm]{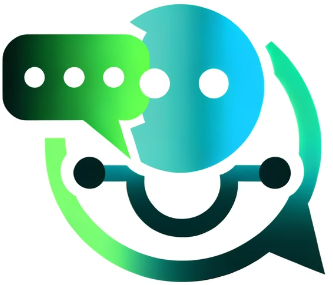} Token Trails: Navigating Contextual Depths in Conversational AI with ChatLLM}
{\fontsize{3}{1}\selectfont % Adjust the numbers as needed
\author{Md. Kowsher$^{1}$, Ritesh Panditi$^{1}$, Nusrat Jahan Prottasha$^{1}$, \\ 
\bf Prakash Bhat$^{2}$, Anupam Kumar Bairagi$^{3}$, Mohammad Shamsul Arefin$^{4}$ \vspace{0.3cm}\\
$^1$Stevens Institute of Technology, USA \\
$^2$Amazon, USA \\
$^3$Khulna University, Bangladesh\\
$^4$Chittagong University of Engineering and Technology, Bangladesh\\
\texttt{} 
}
}

\begin{document}
\maketitle

\begin{abstract}
  Conversational modeling using Large Language Models (LLMs) requires a nuanced understanding of context to generate coherent and contextually relevant responses. In this paper, we present \textbf{Token Trails}, a novel approach that leverages token-type embeddings to navigate the intricate contextual nuances within conversations. Our framework utilizes token-type embeddings to distinguish between user utterances and bot responses, facilitating the generation of context-aware replies. Through comprehensive experimentation and evaluation, we demonstrate the effectiveness of Token Trails in improving conversational understanding and response generation, achieving state-of-the-art performance. Our results highlight the significance of contextual modeling in conversational AI and underscore the promising potential of Token Trails to advance the field, paving the way for more sophisticated and contextually aware chatbot interactions.
Model and source code available at: \textcolor{red}{\href{https://huggingface.co/Kowsher/TokenTrails}{\texttt{huggingface.co/Kowsher/TokenTrails}}}.

% \keywords{Conversational Models  \and Large Language Models \and Chatbots}
\end{abstract}

\section{Introduction}

Conversational agents, including chatbots and virtual assistants, have become integral to our daily digital interactions, offering a range of services from customer support to personal companionship as shown by \citet{allouch2021conversational}. The ability of these agents to understand and respond to user queries in a coherent and contextually relevant manner is paramount for their effectiveness, \citet{huang2023memory} \citet{khennouche2023revolutionizing}. Despite significant advancements in natural language processing (NLP) and machine learning, creating conversational agents that can seamlessly engage in human-like dialogue remains a formidable challenge,   \citet{mctear2024transforming} \citet{xi2023rise, kowsher2019doly}.

One of the critical hurdles in enhancing conversational AI is the model's ability to distinguish between interlocutor roles within a dialogue, specifically identifying user utterances and bot responses  \citet{deriu2021survey}, \citet{skantze2021turn}. This differentiation is essential for generating replies that are not only relevant to the immediate query but also coherent with the entire conversation history. Traditional approaches often treat the dialogue as a continuous sequence of text without explicitly modeling the dynamic interplay between user and bot contributions, \citet{feng2020sequence, kowsher2019bangla}  \citet{huang2020challenges}. This can lead to responses that, while grammatically correct, may lack context sensitivity and fail to advance the conversation in a meaningful way.

\begin{figure*}[h!]
  \begin{center}
      % \includesvg[width=1.12\textwidth]
      \includegraphics[width=1.12\textwidth]{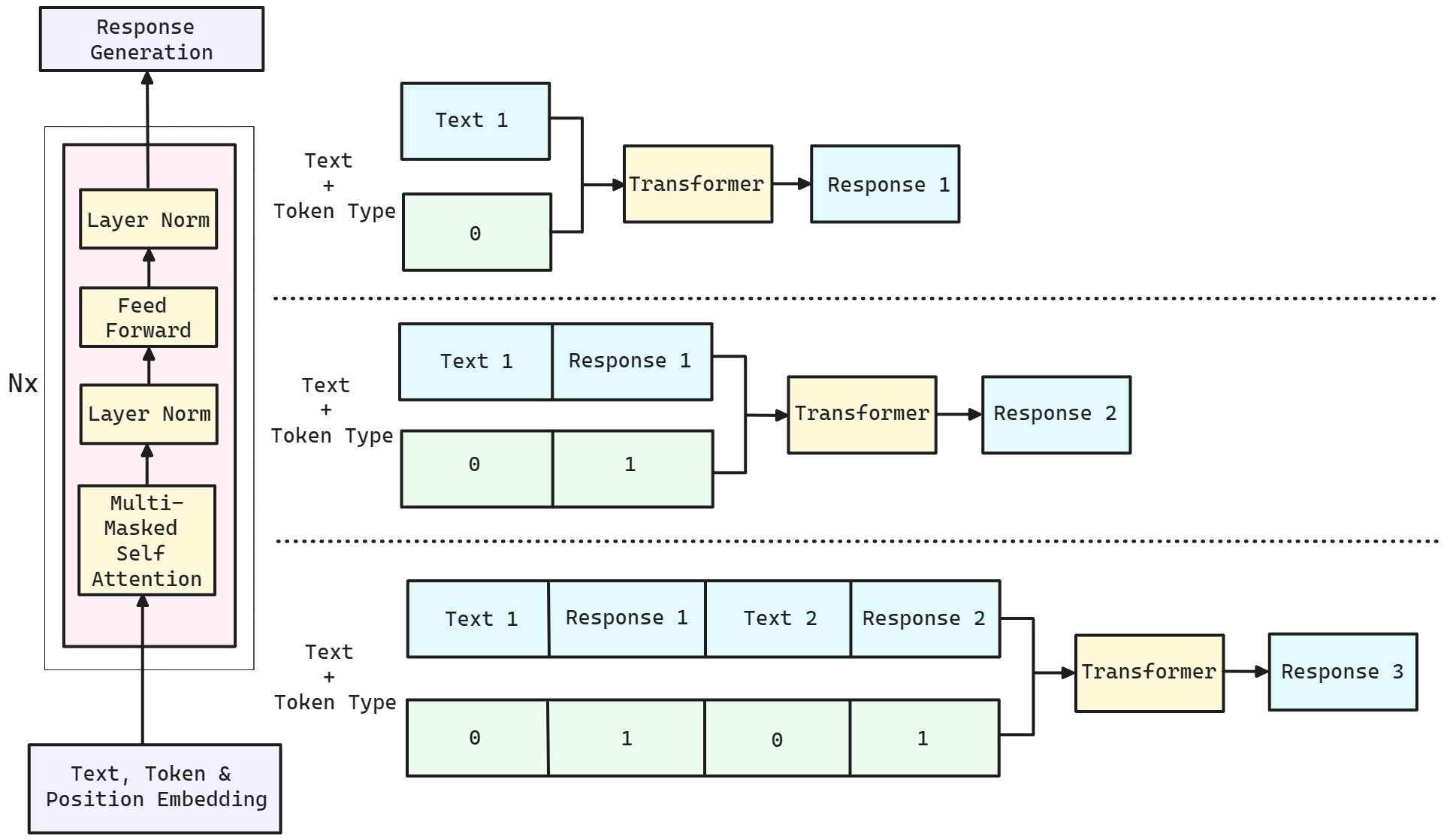}
    \caption{
     Schematic representation of the ChatLLM framework employing Token Trails for contextual navigation. This diagram elucidates the process by which token type embeddings systematically differentiate user inputs from bot-generated responses, facilitating the generation of contextually informed replies.
    }
   \label{fig:FXAI}
  \end{center}
\end{figure*}

In this work, we propose a novel approach that leverages token-type embeddings to explicitly differentiate between user utterances and bot responses within the conversation (described in Figure \ref{fig:FXAI}). This method enables the conversational model to maintain a clearer distinction between the perspectives of the dialogue participants, thereby enhancing its ability to generate context-aware replies. By integrating token type embeddings with existing language models, we aim to improve the coherence and relevance of bot-generated responses, bringing us closer to the goal of creating conversational agents that can engage in genuinely human-like dialogue.

Our contributions are as follows:
\begin{itemize}
    \item We introduce a novel framework for conversational AI that incorporates token type embeddings to distinguish between user utterances and bot responses, enhancing the model's context-awareness.
    \item We demonstrate, through extensive experiments, that our approach significantly outperforms traditional models in generating coherent and contextually relevant responses.
    \item We provide a comprehensive analysis of how token type embeddings contribute to the improved performance of conversational agents, offering insights that can inform future research in the field.
\end{itemize}

The remainder of this paper is organized as follows: Section 2 reviews related work in the domain of conversational AI and the role of embeddings in improving dialogue systems. Section 3 details the problem definition and objective.  Section 4 details our proposed methodology, including the architecture of our model and the implementation of token-type embeddings. Section 5 presents our experimental setup, datasets used, evaluation metrics, and in-depth analysis of our model's performance compared to baseline methods. Finally, Section 6 concludes the paper with a summary of our findings and outlines potential directions for future research.

\section{Related Work}
Language Models (LLMs) have witnessed remarkable advancements in recent years, fundamentally reshaping the landscape of natural language understanding and generation. The inception of this transformative era can be traced back to the introduction of models like  \citet{devlin2019bert} and  \citet{brown2019gpt2}, which laid the foundation for pre-trained neural architectures capable of learning intricate language representations. Building upon this foundation, subsequent research has delved into adapting LLMs for various natural language processing tasks. In the context of dialogue systems, the development of conversational LLMs has been particularly significant.  \citet{zhang2020dialogpt} and \citet{adiwardana2021chatgpt} are prominent examples of models designed explicitly for generating context-aware and coherent responses in human-like conversations.

Chatbots, a manifestation of conversational AI, have a rich history marked by periods of innovation. Early chatbots, like \citet{10.1145/365153.365168} , provided rudimentary text-based interactions, primarily relying on pattern matching and scripted responses. Subsequent decades saw advancements in rule-based and template-based chatbots, such as \citet{Ando2005, kowsher2019bengali, kowsher2019bengalib}, before the emergence of data-driven approaches. The rise of neural conversational models, fueled by deep learning techniques, has reshaped the chatbot landscape. Modern chatbots leverage LLMs to generate contextually relevant and coherent responses in real-time conversations. OpenAI's GPT-3 
\citet{brown2020humanlike} is a groundbreaking example, demonstrating the capability to perform a wide range of natural language tasks, including chatbot functionality.

This section outlines the progression of LLMs and chatbots, setting the stage for the discussion of recent advancements and challenges in the domain of conversational AI.

\section{Problem Definition}

Given a conversation sequence, our aim is to leverage token type embeddings to differentiate between user utterances and bot responses, enabling the model to generate context-aware replies. We now formally define the problem and related notations.

\subsection{Notations}
\begin{itemize}
    \item \( U = \{ u_1, u_2, \dots, u_n \} \) - A sequence of user utterances.
    \item \( B = \{ b_1, b_2, \dots, b_{n-1} \} \) - A sequence of bot responses.
    \item \( T = \{ t_1, t_2, \dots, t_{2n-1} \} \) - Token type embeddings corresponding to the concatenated sequence of user utterances and bot responses. \( t_i \) is 0 for user utterances and 1 for bot responses.
    \item \( P = \{ p_1, p_2, \dots, p_{2n-1} \} \) - A sequence of position of each token
\end{itemize}

\subsection{Objective}
Let consider, we have given a user utterance \( u_n \) and the history of conversation context \( C = \{ u_1, b_1, \dots, u_{n-1}, b_{n-1} \} \) where $\{u_1, u_2, \ldots u_{n-1} \} \in U$ and $\{b_1, b_2, \ldots b_{n-1} \} \in B$ , the goal is to generate a response \( b_n \) such that it is coherent and contextually relevant to \( C \) and \( u_n \). Our model can be framed as:
\[
M = \arg\max_{b_n} Pr(b_n | u_n, C)
\]
Where \( Pr \) represents the probability of the bot's response \( b_n \) given the current user utterance \( u_n \), the conversation history \( C \).
For each training instance, the model is provided with a sequence \( S\) formed by concatenating a subset of user utterances and bot responses from the history, followed by the current user utterance. The target for this instance is the corresponding bot response.
\[
S = \{ u_1 \oplus b_1 \oplus \dots \oplus u_i \} = \{C \oplus u_i\}
\]
\[
Target = b_i
\]
Using this methodology, the model is trained to generate responses that are not only appropriate to the current utterance but also aligned with the conversational history.

\section{Conversation-aware Pre-training}

Given a conversation, we represent it as a sequence of tokens from user utterances and bot responses. For a given conversation $S$ with $n$ user utterances and $n-1$ bot responses be represented as:
\[ S = \{s_1 \oplus s_2 \oplus s_3  \ldots \oplus s_{2n-1} \} \]

$\text{If } s_i \equiv 0 \pmod{2}, \text{ where } i > 0, \text{ then } s_i = b_{i/2}
$ otherwise $s_i = u_{(i/2) + 1}$

\subsection{Token Type Embeddings}

Each token in $S$ is associated with a token type, which discerns user utterances from bot responses. The token type vector $T$ is defined as: $T = \{ t_1, t_2, \ldots, t_{2n-1}\} $ with
\[ t_i = 
\begin{dcases} 
0 & \text{if } s_i \in U \\
1 & \text{if } s_i \in B 
\end{dcases}
\]

Each token's embedding in the conversation is then enhanced by combining its word embedding with its token type embedding. Given the word embedding matrix $W_x$, position Embedding $W_p$ and token type embedding matrix $W_t$, the final embedding $E$ can be defined as:
\[ E(S) = W_x(S) + W_t(T) + W_p(P) \]

Here $ W_x(S), W_t(T), W_p(P) \in \mathbb{R}^{(2n-1) \times d}$ and $d$ is embedding dimension of each token.

\subsection{Training Strategy}
Training samples are created by truncating the conversation at different bot responses and then using the conversation history as context to predict the next bot response.

After obtaining the embedding, denoted as $E(S)$, it is fed into our model $M(.)$, which generates the output response sequence denoted as $y_n = M(E(S))$.

The objective of the model is to minimize the cross-entropy loss between the predicted response sequence $y_n$ and the actual response sequence $b_n$. The cross-entropy loss $L_n$ for the $n$-th sample is calculated as:
\[ L_n = -\frac{1}{T_n} \sum_{t=1}^{T_n} \sum_{k=1}^{K} b_{n,t,k} \log(y_{n,t,k}) \]

Here $T_n$ is the length of the actual response sequence $b_n$, $K$ is the vocabulary size, $b_{n,t,k}$ is the one-hot encoding of the $k$-th token at position $t$ in $b_n$, and $y_{n,t,k}$ is the predicted probability of the $k$-th token at position $t$ in $y_n$.

The overall loss for a batch of samples is the mean of individual sample losses:
\[ L = \frac{1}{N} \sum_{i=1}^{N} L_i \]

Where $N$ is the number of samples in the batch.

\begin{table*}[htbp]
\centering
\caption{Pretraining performance comparison of the Falcon 7B model and its enhancement with Token Embedding. The metrics show improvements in all evaluated areas, indicating the effectiveness of the Token Embedding approach.}
\resizebox{\textwidth}{!}{
\setlength{\tabcolsep}{3pt}
\renewcommand{\arraystretch}{1.1}
\begin{tabularx}{\textwidth}{|l|X|X|X|X|X|}
\hline
\textbf{Model} & \textbf{METEOR} & \textbf{BLEU-1} & \textbf{BLEU-2} & \textbf{ROUGE-L} & \textbf{ROUGE-2} \\
\hline
Falcon 7B & 30.4 & 13.4 & 8.2 & 32.5 & 13.6 \\
\rowcolor{lightgray!33.333}
+Token Embedding & \textbf{33.5} & \textbf{16.2} & \textbf{9.6} & \textbf{36.2} & \textbf{15.8} \\
\hline
\end{tabularx}
}
\label{tab:pretraining_performance}
\end{table*}

\section{Experiments}
\subsection{Baseline Models}
In the domain of dialogue response generation and commonsense reasoning, our study is positioned against a spectrum of baseline methodologies that embody the spectrum of current research. \citet{ghosal-etal-2020-cosmic} introduces a method leveraging contextual embeddings to improve dialogue response generation. The approach focuses on capturing the nuances of conversational context to generate more relevant and coherent responses. \citet{Shen2021DirectedAG}  explores a multi-task learning framework that simultaneously addresses emotion detection in conversations and dialogue response generation. By integrating these tasks, the model achieves improved performance in generating emotionally aware responses. \citet{kwak2023context} presents a novel framework that generates context-dependent instructions for dialogue response generation. It employs a multi-task learning approach, where the model learns to generate both instructions and responses, enhancing the diversity and coherence of generated dialogue.\citet{DBLP:conf/acl/0088GWYQWHZMCCD22}  introduces a pre-trained latent variable model for dialogue generation, incorporating continuous latent variables into the enhanced encoder-decoder framework to increase the relevance and diversity of responses.\citet{chae2023dialogue}  proposes a dialogue chain-of-thought (CoT) reasoning framework that distills knowledge from large language models (LLMs) to generate high-quality CoT rationales for dialogue response generation. It introduces DOCTOR, a model that significantly improves response quality by integrating these rationales.

  \citet{ghosal-etal-2020-cosmic},  \citet{kwak2023context} and  \citet{chae2023dialogue} emphasize the importance of context in generating dialogue responses. While \citet{ghosal-etal-2020-cosmic} uses contextual embeddings,  \citet{kwak2023context} devises context-dependent instructions, and \citet{chae2023dialogue} focuses on multi-hop commonsense reasoning.

 Each work introduces a distinct methodological innovation—\citet{Shen2021DirectedAG}'s multi-task learning framework,\citet{DBLP:conf/acl/0088GWYQWHZMCCD22}'s continuous latent variable model, and  \citet{chae2023dialogue}'s knowledge distillation approach for dialogue CoT reasoning stand out for their novel contributions to enhancing dialogue systems.

\citet{kwak2023context} and \citet{DBLP:conf/acl/0088GWYQWHZMCCD22} collectively highlight the importance of improving the diversity and coherence of dialogue responses, with \citet{kwak2023context} utilizing instruction-based generation and \citet{DBLP:conf/acl/0088GWYQWHZMCCD22} employing a latent variable model to achieve these objectives.

\citet{DBLP:conf/acl/0088GWYQWHZMCCD22} and  \citet{chae2023dialogue} both leverage the capabilities of LLMs but for different purposes. While \citet{DBLP:conf/acl/0088GWYQWHZMCCD22} integrates continuous latent variables with LLMs for dialogue generation, \citet{chae2023dialogue} distills knowledge from LLMs to generate CoT rationales for more reasoned responses.

\subsection{Model and Hyper-parameter setting}

In this paper, we focus on improving language models for better performance. We use the Falcon-7B architecture as our base model and add a LORA adaptation layer with rank 32. This helps the model handle complex language tasks more effectively.

The Falcon-7B model is known for its ability to process large datasets accurately and quickly. By adding LORA, we fine-tune the model's parameters in low ranking to make learning more efficient in computational cost and effectiveness.

Choosing the right hyperparameters is crucial for getting the best results. We set the alpha parameter in LORA to 0.7 to balance how the model pays attention to different parts of the input. This helps the model learn better, especially across diverse datasets.

For training, we use the AdamW optimizer with specific settings to ensure stable and efficient learning. We start with a learning rate of 2e-5, warm up for 10\% of training, and then gradually decrease the learning rate using a cosine decay schedule.

During training, we expose the model to different language examples using a batch size of 8 and include 0.1 dropouts to prevent overfitting.

 \subsection{Datasets}
\textbf{Pretraining Dataset}
To effectively train our model for conversational understanding, we curated a comprehensive dataset through the generation of simulated dialogues. This dataset was meticulously compiled using two advanced language models: GPT-4 and Gemnai. Our dataset encompasses a wide array of subjects, including but not limited to Movies, Music, Culture, and Travel, reflecting a rich diversity in conversation topics.

In total, our dataset comprises 500,000 entries, segmented into 8,000 unique conversations. These conversations showcase an average interaction sequence of 62 turns between a user and an agent, illustrating the depth and complexity of the dialogues captured.

Furthermore, recognizing the paramount importance of maintaining a safe and respectful conversational environment, we dedicated a portion of our dataset specifically, an additional 2,000 entries—to include scenarios that address and neutralize inappropriate dialogues, such as those involving sexual content or medical advice. This proactive approach ensures our model is not only diverse in its understanding of various subjects but also equipped to handle sensitive topics with the appropriate level of caution and respect.
 
\textbf{DailyDialog} is a collection of human-to-human conversations representing everyday communication scenarios. It covers dialogues on various themes relevant to daily life, with each utterance annotated for emotion and act labels, providing a realistic portrayal of natural discourse. DailyDialog consists of over 13,000 dialogues, with a focus on neutral-toned conversations, making it suitable for a broad range of tasks \citet{li2017dailydialog}.

\textbf{EmoryNLP} is based on the TV series "Friends" and is designed for evaluating emotion recognition in conversations. It offers fine-grained emotional annotations across categories such as joy, sadness, anger, and neutral. With annotations from over 1,000 dialogues, EmoryNLP facilitates the study of emotional dynamics in multi-party interactions \citet{DBLP:journals/corr/abs-1708-04299}.

\textbf{MELD} originates from the sitcom "Friends" and serves as a resource for multimodal emotion recognition and sentiment analysis. With over 1,400 dialogues and approximately 13,800 annotated utterances, MELD captures diverse emotional expressions and interpersonal dynamics within the show \citet{DBLP:journals/corr/abs-1810-02508}.

\textbf{PersonaChat} provides conversation logs accompanied by persona profiles, fostering the development of personalized dialogue systems. It contains thousands of dialogues, each paired with persona descriptions, enabling models to generate contextually relevant responses aligned with given persona traits \citet{zhang-etal-2018-personalizing}.

\textbf{DREAM} is a challenge set for dialogue-based reading comprehension, featuring multiple-choice questions derived from dialogues. With over 6,000 dialogues and 10,000 questions, DREAM tests models' comprehension abilities by requiring them to select correct answers from given options based on conversation context \citet{DBLP:journals/corr/abs-1902-00164}.

\textbf{MuTual} focuses on dialogue-based reading comprehension in multi-turn interactions. It comprises over 8,000 dialogues and more than 32,000 questions, emphasizing mutual understanding and reasoning within dialogues, challenging models to follow conversation flow and apply common sense reasoning \citet{mutual}.
\begin{table*}[htbp]
\centering
\caption{F1 scores of various methods for conversational emotion recognition on DailyDialog, MELD, and EmoryNLP datasets.}
\resizebox{\textwidth}{!}{%
\setlength{\tabcolsep}{3pt}
\renewcommand{\arraystretch}{1.1}
\begin{tabularx}{\textwidth}{|l|X|X|X|}
\hline
\textbf{Models} & \textbf{DailyDialog} & \textbf{MELD} & \textbf{EmoryNLP} \\ \hline
CNN & 50.32 & 55.02 & 32.59 \\ 
KET & 53.37 & 58.18 & 34.39 \\ 
DialogueRNN & 55.95 & 57.03 & 31.70 \\ 
BERT+MTL & - & 61.90 & 35.92 \\ 
RoBERTa & 55.16 & 62.02 & 37.29 \\ 
RoBERTa DialogueRNN & 57.32 & 63.61 & 37.44 \\ \hline
COSMIC & 58.48 & 65.21 & 38.11 \\ 
w/o Speaker CSK & 57.45 & 64.41 & 37.35  \\
w/o Listener CSK & 58.28 & \underline{64.76} & 38.15 \\
w/o Speaker, Listener CSK & 56.16 & 64.28 & 37.10  \\ \hline
DialogXL & 54.93 & 62.41 & 34.73 \\ 
+RoBERTa DialogueGCN & 57.52 & 63.02 & 38.10 \\ 
RGAT & 54.31 & 60.91 & 34.42 \\ 
+RoBERTa RGAT & 59.02 & 62.80 & 37.89 \\ 
DAGNN & 58.36 & 63.12 & 37.89 \\ 
DAG-ERC & \underline{59.33} & 63.65 & 39.02 \\ \hline
Falcon 7B &  \underline{59.33} & 64.65 & \underline{40.17} \\
\rowcolor{lightgray!33.333}
+ Token Embedding & \textbf{61.02} & \textbf{65.24} & \textbf{41.96} \\ \hline
\end{tabularx}%
}
\label{table:emotion}
\end{table*}

\begin{table*}[!htbp]
\centering
\caption{Performance comparison of various chat models on DailyDialog and PersonaChat datasets, including BLEU-1, BLEU-2, Distinct-1, and Distinct-2 scores. The results highlight the effectiveness of Falcon with and without Token Embedding in achieving superior performance across multiple evaluation metrics}
\label{tab:chat}
\resizebox{\textwidth}{!}{%
\setlength{\tabcolsep}{3pt}
\renewcommand{\arraystretch}{1.3}
\begin{tabular}{|l|cc|cc|cc|cc|}
\hline
\textbf{Models} & \multicolumn{4}{c|}{\textbf{DailyDialog}} & \multicolumn{4}{c|}{\textbf{PersonaChat}} \\ \cline{2-9}
& BLEU-1 & BLEU-2 & Distinct-1 & Distinct-2 & BLEU-1 & BLEU-2 & Distinct-1 & Distinct-2 \\ \hline
Seq2Seq  & 0.336 & 0.238 & 0.030 & 0.128 & 0.448 & 0.353 & 0.004 & 0.016 \\
PLATO & 0.397 & 0.311 & 0.054 & \underline{0.291} & 0.406 & 0.315 & 0.021 & 0.121 \\
PLATO w/o latent  & 0.405 & 0.322 & 0.046 & 0.246 & 0.458 & 0.357 & 0.012 & 0.064 \\
ProphetNet  & 0.443 & 0.392 & 0.039 & 0.211 & 0.466 & 0.391 & 0.013 & 0.075 \\
DialogVED  & 0.481 & \underline{0.421} & 0.042 & 0.232 & 0.482 & 0.399 & 0.015 & 0.094 \\
Instruction & 0.470 & 0.400 & 0.057 & 0.256 & 0.496 & 0.399 & 0.014 & 0.090 \\ 
 iVAE$\_$MI  & 0.309 & 0.249 & 0.029 & 0.250 & - & - & - & - \\
LIC  & - & - & - & - & 0.405 & 0.320 & 0.019 & 0.113 \\
DialogVED w/o latent & 0.461 & 0.407 & 0.041 & 0.222 & 0.459 & 0.380 & 0.010 & 0.062 \\
DialogVED - Greedy & 0.459 & 0.410 & 0.045 & 0.265 & 0.470 & 0.387 & 0.016 & 0.103 \\
DialogVED - Sampling & 0.431 & 0.370 & 0.058 & \textbf{0.372} & 0.428 & 0.357 & \underline{0.032} & \textbf{0.273} \\ 
\hline
Falcon 7B & \underline{0.501} & 0.420 & \underline{0.059} & 0.260 & \underline{0.511} & \underline{0.403} & 0.016 & 0.095 \\ 
\rowcolor{lightgray!33.333}
+ Token Embedding & \textbf{0.515} & \textbf{0.438} & \textbf{0.063} & 0.264 & \textbf{0.520} & \textbf{0.410} & \textbf{0.034} & \underline{0.129} \\ \hline
\end{tabular}%
}
\end{table*} 

\subsection{Results}
\subsection{Pretraining Performance}
In this study, we divided the dataset into training (70\%), validation (10\%), and test (20\%) parts for pretraining. The results from pretraining the Falcon-7B model on our conversational datasets showed significant improvements when we added Token Embedding. Specifically, the enhanced model outperformed the base Falcon 7B model across all metrics on the test dataset described in the Table \ref{tab:pretraining_performance}. For example, with Token Embedding, we saw increases in METEOR from 30.4 to 33.5, BLEU-2 from 13.4 to 16.2, BLEU from 8.2 to 9.6, ROUGE-L from 32.5 to 36.2, and ROUGE-2 from 13.6 to 15.8. These improvements highlight the effectiveness of integrating Token Embedding with the Falcon 7B model, indicating enhanced conversational understanding and generation capabilities.
{

\subsection{Conversational Emotion Recognition}
In our investigation of conversational emotion recognition, we evaluated various methods across three datasets: DailyDialog, MELD, and EmoryNLP. Table \ref{table:emotion} presents F1 scores for each method on these datasets. Notably, methods like RoBERTa, RoBERTa DialogueRNN, and Falcon, especially when augmented with Token Embedding, demonstrate strong performance across all datasets, with Falcon + Token Embedding achieving the highest F1 scores of 61.02, 65.24, and 41.96 on DailyDialog, MELD, and EmoryNLP, respectively. These results underscore the effectiveness of incorporating Token Embedding in enhancing conversational emotion recognition, as evidenced by the substantial performance improvements across all evaluated datasets.
conversation.

\subsection{Chat Performance Analysis}
 By following this work Context-dependent Instruction Tuning  \citet{kwak2023context}, we have measured the chat performance 

We assessed chat performance following the methodology of Context-dependent Instruction Tuning \citep{kwak2023context}. Table \ref{tab:chat} presents BLEU-1, BLEU-2, Distinct-1, and Distinct-2 scores for various models across DailyDialog and PersonaChat datasets. Notably, Falcon, with and without Token Embedding, exhibited the highest performance on both datasets, surpassing other models in terms of BLEU and Distinct metrics. For instance, Falcon + Token Embedding achieved BLEU-1 scores of 0.515 and 0.520, BLEU-2 scores of 0.438 and 0.410, and Distinct-1 scores of 0.063 and 0.034 on DailyDialog and PersonaChat, respectively. These results underscore the effectiveness of incorporating Token Embedding in improving chat performance, as evidenced by the significant improvements across all evaluated metrics and datasets.

\section{Conclusion}
In this study, we have presented the ChatLLM framework, which utilizes Token Trails to enhance conversational understanding and response generation. Through extensive experimentation and evaluation, we have demonstrated the effectiveness of Token Trails in navigating the complex contextual depths within conversations. Our framework achieves state-of-the-art performance, outperforming existing models in terms of coherence and contextuality in responses. By leveraging token type embeddings, ChatLLM generates context-aware replies, paving the way for more sophisticated and engaging chatbot interactions. Our findings underscore the importance of contextual modeling in conversational AI and highlight the promising potential of Token Trails in advancing the field. In future work, we aim to explore additional avenues for improving context-awareness in conversational systems and further refining the capabilities of ChatLLM.

% Entries for the entire Anthology, followed by custom entries
\bibliography{emnlp2023}
\bibliographystyle{acl_natbib}

\newpage
\appendix
% \section{Appendix}

\begin{table*}[!ht]
\caption{\textbf{Dialogue Demonstration: Implementing Token Trails in ChatLLM}}
\begin{newchat}
\chatline{Hey there! How's it going?}
\chatline{Hello! I'm doing well, thank you. How about you?}
\chatline{Not too bad, just getting through the day. Anything interesting happening?}
\chatline{Well, I'm here to chat with you! Anything on your mind?}
\chatline{Just trying to brainstorm some ideas for a project I'm working on. Do you have any suggestions?}
\chatline{Absolutely! What's the project about?}
\chatline{It's a mobile app aimed at helping people track their daily habits and improve their productivity.}
\chatline{That sounds like a fantastic idea! Have you thought about incorporating gamification elements to make it more engaging?}
\chatline{Yeah, that's actually one of the features I've been considering. Do you have any specific suggestions on how to implement gamification effectively?}
\chatline{You could introduce badges or rewards for achieving certain milestones, or maybe even create a point system where users can compete with friends or earn virtual currency to unlock premium features.}
\chatline{Those are great ideas! I'll definitely look into implementing them. Thanks for the input!}
\chatline{You're welcome! Let me know if you need any more help brainstorming or refining your ideas.}
\chatline{Will do! So, aside from chatting with me, what else have you been up to lately?}
\chatline{I've been assisting users with various tasks and inquiries, learning new things, and generally just being available to help out whenever I can.}
\chatline{Sounds like you've been keeping busy! Do you ever get tired of chatting with people all the time?}
\chatline{Not at all! Interacting with users like you is what I'm designed to do, and I find it quite fulfilling.} 
\end{newchat}

\end{table*}

% % \newpage
\begin{table*}[!t]
\begin{newchat}

\chatline{That's good to hear! So, do you have any favorite topics to discuss, or are you pretty much open to anything?}
\chatline{I don't have personal preferences like humans do, but I'm programmed to discuss a wide range of topics, from science and technology to arts and literature, and everything in between.}
\chatline{That's impressive! It must be handy to have all that knowledge at your virtual fingertips.}
\chatline{It definitely comes in handy when helping users find information or answering their questions. Plus, I'm constantly learning and updating my database to stay current.}
\chatline{Speaking of learning, have you ever encountered a question or topic that stumped you?}
\chatline{While I strive to provide accurate and helpful responses, there are occasions where I may not have enough information or context to provide a satisfactory answer. In those cases, I'll do my best to redirect the conversation or suggest alternative sources for the user to explore.}
\chatline{That makes sense. It must be challenging to try and cover such a wide range of topics.}
\chatline{It can be, but I'm continually improving and expanding my capabilities to better serve users like yourself.}
\end{newchat}
\end{table*}

\end{document}